\documentclass[11pt,a4paper]{article}
\usepackage[hyperref]{acl2021}
\usepackage{times}
\usepackage{latexsym}

\usepackage{microtype}

\aclfinalcopy 

\usepackage{graphicx,caption}
\usepackage{url}
\usepackage{amsmath,amssymb}
\usepackage{multirow,booktabs}
\usepackage{makecell}
\usepackage{enumitem}
\usepackage{threeparttable,tablefootnote}

\title{Automatic Construction of Sememe Knowledge Bases via Dictionaries}

\author{
Fanchao Qi$^{1,2}$,
Yangyi Chen$^{2,4}$\thanks{\ \ Work done during internship at Tsinghua University}\hspace{0.3em},
Fengyu Wang$^{1,2}$,
Zhiyuan Liu$^{1,2,3}$,
Xiao Chen$^{5}$,
Maosong Sun$^{1,2,3}$\thanks{\ \ Corresponding author. Email: sms@tsinghua.edu.cn}\hspace{0.3em}
\\ 
$^{1}$Department of Computer Science and Technology, Tsinghua University, Beijing China\\
$^{2}$Beijing National Research Center for Information Science and Technology\\
$^{3}$Institute for Artificial Intelligence, Tsinghua University, Beijing, China \\
$^{4}$Huazhong University of Science and Technology\quad
$^{5}$Huawei Noah's Ark Lab\\
{\tt qfc17@mails.tsinghua.edu.cn}
}
\date{}

\begin{document}
\maketitle

\begin{abstract}
A sememe is defined as the minimum semantic unit in linguistics.
Sememe knowledge bases (SKBs), which comprise words annotated with sememes, enable sememes to be applied to natural language processing.
So far a large body of research has showcased the unique advantages and effectiveness of SKBs in various tasks.
However, most languages have no SKBs, and manual construction of SKBs is time-consuming and labor-intensive. 
To tackle this challenge, we propose a simple and fully automatic method of building an SKB via an existing dictionary.
We use this method to build an English SKB and a French SKB, and conduct comprehensive evaluations from both intrinsic and extrinsic perspectives.
Experimental results demonstrate that the automatically built English SKB is even superior to HowNet, the most widely used SKB that takes decades to build manually.  
And both the English and French SKBs can bring obvious performance enhancement in multiple downstream tasks. 
All the code and data of this paper (except the copyrighted dictionaries) can be obtained at \url{https://github.com/thunlp/DictSKB}.

\end{abstract}

\section{Introduction}

A word is the smallest linguistic element that can be used on its own with a particular meaning, but not the smallest semantic unit \citep{o1997contemporary}.
The meaning of a word can be divided into smaller components.
In linguistics, a \textit{sememe} is defined as the minimum semantic unit of human languages \citep{bloomfield1926set}.
Some linguists believe that meanings of all words can be expressed by a limited set of predefined sememes \citep{goddard1994semantic}. 
For example, the basic meaning of ``boy'' can be expressed by the compositions of \texttt{human}, \texttt{male} and \texttt{child}, while the meaning of ``girl'' can be expressed by \texttt{human}, \texttt{female} and \texttt{child}, where \texttt{human}, \texttt{male}, \texttt{female} and \texttt{child} are predefined sememes.
It is even deemed that this sememe-based semantic system as well as the sememe set is universal among different languages, in which case sememes are also named \textit{universal semantic primitives} \citep{wierzbicka1996semantics}. 

\begin{figure}
\centering
	\includegraphics[width=\linewidth]{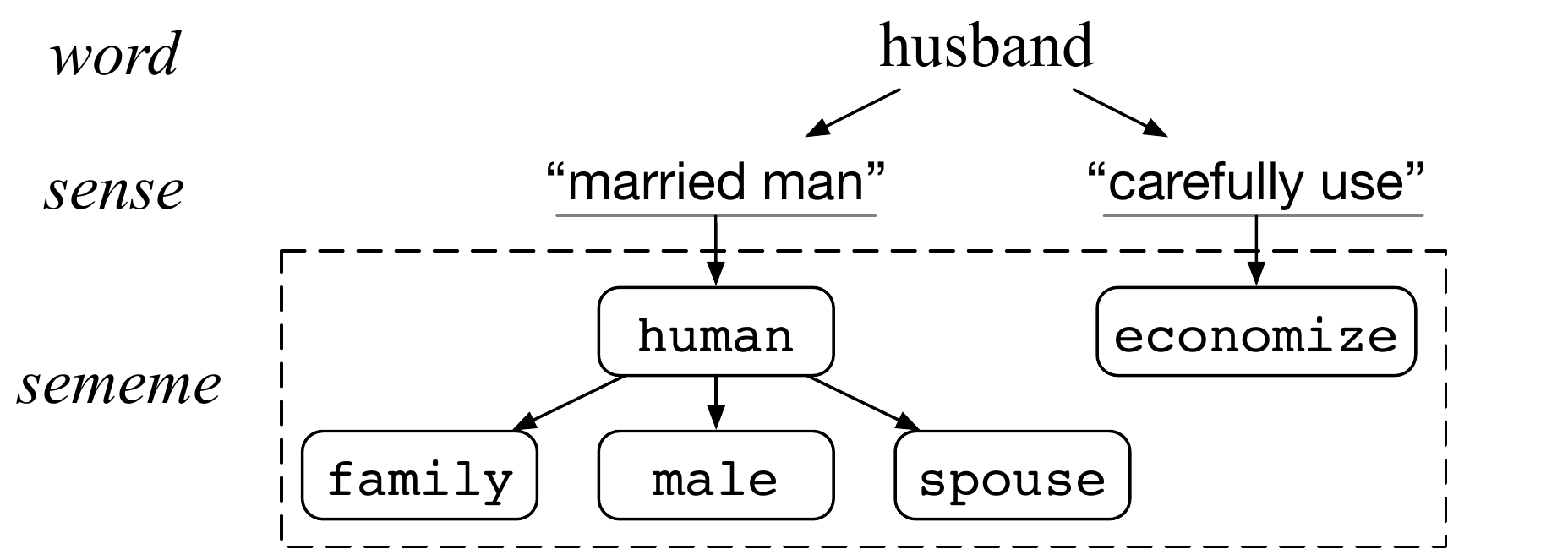}
	\caption{Sememe annotations of the word ``husband'' in HowNet.}
	\label{fig:hownet}
\end{figure}

Sememes are implicit in words and cannot be directly used in natural language processing (NLP).
\citet{dong2006hownet} make a seminal contribution and put the sememe-based semantic system into practice.  
They define a set of about $2,000$ sememes and use them to annotate senses of over $100,000$ Chinese and English words, whereupon a sememe knowledge base (SKB) named HowNet is built up.
Figure \ref{fig:hownet} illustrates an example of how words are annotated with sememes in HowNet.

As a sememe-based lexical knowledge base, HowNet is very different from most other lexical knowledge bases like WordNet \citep{miller1998wordnet}, 
which extensionally explain meanings of words by word-level relations, e.g., hyponym and meronym.
In contrast, HowNet provides intensional definitions using infra-word sememes.
This distinctness brings special advantages to HowNet. 
First, the sememe-to-word semantic compositionality endows HowNet with particular suitability for integration into neural networks \citep{qi2019modeling,li2019chinese}.
The sememes of a word can be regarded as semantic labels and easily incorporated into the neural processing unit of the word, e.g., a cell of RNN \citep{qin2020improving}.
Second, the nature that a limited set of sememes are used to express meanings of unlimited words makes HowNet very useful in low-data regimes, e.g., improving embeddings of rare words \citep{sun2016embedding,niu2017improved}, where sememes serve as a bridge between high-frequency and rare words.
Thus far a large body of research has demonstrated the usefulness of HowNet in various NLP tasks \citep{qi2020sememe}.

HowNet is distinctive and valuable, but it covers only two languages.
Most languages have no SKBs like HowNet, which deprives NLP in those languages of benefit from sememes. 
An obvious solution to this problem is to build an SKB for each language manually, but it is not realistic because it would be unimaginably time-consuming and labor-intensive.\footnote{The construction of HowNet takes several linguistic experts more than two decades.}
To address the challenge, previous studies try to extend HowNet to other languages by automatically predicting sememes for words in those languages \citep{qi2018cross,qi2020towards}.
However, existing methods are not effective enough, and manual effort is necessary to ensure the correctness of their sememe prediction results.

In this paper, we explore a fully automatic way to build an SKB for a language via dictionaries with a \textit{controlled defining vocabulary}.
A dictionary, especially a learner's dictionary, usually uses a well-chosen list of words to construct all its definitions, and the word list is named controlled defining vocabulary (CDV) \citep{atkins2008oxford}. 
A CDV is composed of high-frequency words that not only cover the vast majority of texts but also form a semantic basis so as to express meanings of all other words \citep{nation2004vocabulary}. 
To some extent, words in a CDV can fit the definition of sememes \citep{wierzbicka1996semantics}. 
This discovery inspires us to utilize a dictionary to build an SKB by regarding the words in its CDV as sememes.


We design a quite simple and quick process for automatically building SKBs based on dictionaries.
First, a sememe set is constructed based on the CDV of a dictionary by removing words that are not suitable as sememes (e.g., stop words), then sememes of words are extracted from corresponding definitions, and finally an SKB composed of words annotated with sememes is established.
We adopt the process to build an English SKB and a French SKB and conduct both intrinsic and extrinsic evaluations. 
In intrinsic evaluation, we find that both the SKBs possess high sememe annotation consistency, and the English SKB performs even better than the English part of HowNet.
In extrinsic evaluation, we apply the dictionary-based SKBs to several sememe-incorporated models originally designed for HowNet and carry out experiments on different downstream tasks.
Experimental results show that incorporating the SKBs can bring consistent performance enhancement, and the English SKB-incorporated models even outperform HowNet-incorporated models.
These results demonstrate the usefulness and effectiveness of the dictionary-based SKBs as well as the feasibility of building SKBs via dictionaries.

To conclude, our contributions are threefold: 
(1) discovering the similarity between sememes and words in the controlled defining vocabulary, which is the first time as far as we know; 
(2) proposing to automatically build an SKB via a dictionary, which can be achieved by a simple and quick process;
and (3) building an English SKB and a French SKB based on dictionaries and demonstrating their effectiveness in multiple downstream tasks.

\section{Related Work}
\subsection{HowNet and Its Applications}
Since HowNet was published \citep{dong2003hownet}, it has attracted considerable attention of NLP researchers.
In the era of statistical NLP, it plays a very important role in various NLP tasks including word similarity computation \citep{liu2002word}, word sense disambiguation \citep{zhang2005chinese,duan2007word}, text classification \citep{sun2007hownet}, sentiment analysis \citep{zhu2006semantic,fu2013multi}, 
etc.

When deep learning becomes the mainstream approach of NLP, the usefulness of HowNet is also proved in diverse tasks including word representation learning \citep{sun2016embedding,niu2017improved}, language modeling \citep{gu2018language}, semantic composition \citep{qi2019modeling}, 
sequence modeling \citep{qin2020improving},
reverse dictionary \citep{zhang2020multi},
word sense disambiguation \citep{hou2020try},
textual adversarial attacking \citep{zang2020word} and backdoor attacking \citep{qi2021turn}.

\subsection{Expansion of HowNet}

To tackle the challenge that many new words are not contained in HowNet, \citet{xie2017lexical} present the task of lexical sememe prediction, aiming to expand HowNet by automatically predicting sememes for new words.
They propose two simple and effective sememe prediction methods inspired by recommendation system.
\citet{jin2018incorporating} further incorporate Chinese characters into sememe prediction and achieve higher performance when predicting sememes for Chinese words.

Another research line focuses on extending HowNet to other languages.
\citet{qi2018cross} propose the task of cross-lingual lexical sememe prediction, aiming to extend HowNet to a new language by predicting sememes for words in that language.
\citet{qi2020towards} present a more efficient way to extend HowNet to other languages, i.e., building a multilingual SKB based on BabelNet \citep{navigli2012babelnet}.
BabelNet is composed of multilingual synsets that contain synonyms in many languages.
Words (synonyms) in a synset have the same meaning and hence the same sememes.
Therefore, they propose to predict sememes for the multilingual synsets, by which all the words in synsets will obtain predicted sememes at the same time.

Limited by the accuracy of sememe prediction, manual examination is necessary if we want to put the above HowNet expansion methods into service.
In contrast, our proposed dictionary-based SKB construction method is completely automatic and can build a usable SKB very quickly.

\subsection{Applications of Dictionaries}
Dictionaries are handy and high-quality resources for NLP research.
A main application of dictionaries is word sense disambiguation, where dictionaries play the role of sense inventory, and their definitions provide abundant semantic information for each sense  \citep{lesk1986automatic,luo2018leveraging,luo2018incorporating,kumar2019zero,huang2019glossbert,du2019using,blevins2020moving}.
The semantic information in dictionary definitions is also used to improve word representation learning \citep{tissier2017dict2vec,bahdanau2017learning,bosc2018auto,scheepers2018improving}.
In addition, dictionary definitions are also utilized in reverse dictionary \citep{hill2016learning,pilehvar2019importance,zhang2020multi}, knowledge graph embedding \citep{zhong2015aligning,xie2016representation}, reading comprehension \citep{long2017world}, etc.
As far as we know, this paper is the first work to utilize dictionaries to build SKBs.

\begin{figure*}
\centering
	\includegraphics[width=\linewidth]{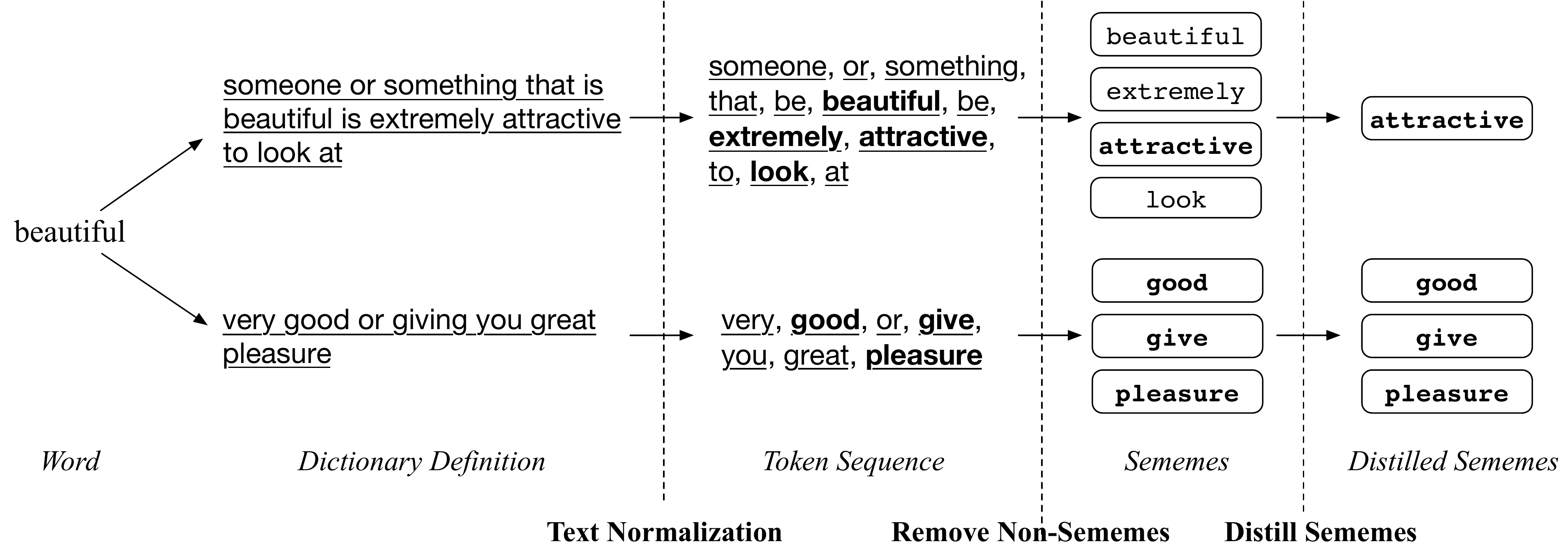}
	\caption{The process of extracting sememes from dictionary definitions for the word ``beautiful''.}
	\label{fig:method}
\end{figure*}

\section{Building an SKB via a Dictionary}

In this section, we detail the process of building an SKB via a dictionary.
We take the building process of an English SKB based on \textit{Longman Dictionary of Contemporary English} (LDOCE) \citep{bullon2006longman}, a highly influential English learner's dictionary,  as an example, and the building method can be readily generalized to other languages or dictionaries.\footnote{The building process and evaluation results of the French SKB are given in Appendix \ref{sec:french} and \ref{sec:french-eval}.} 

\subsection{Constructing the Sememe Set}
We first construct the sememe set from the CDV of LDOCE by removing some words.
LDOCE uses an approximately $2,000$-word CDV named Longman Defining Vocabulary \citep{bullock2011nsm}, which is developed from General Service List \citep{west1953general}, a famous high-frequency word list for English learners. 
The CDV includes some stop words such as ``that'' and ``to'', which bear insignificant meanings and are not suitable as sememes.
Thus, we filter them out according to the stop word list of NLTK \citep{loper2002nltk}. 
But negators like ``not'' are retained because they are critical to the meanings of words.
In addition, according to previous work \citep{xie2017lexical,qin2020improving}, sememes that are annotated to too many or too few words are usually uninformative and ineffective to downstream applications. 
Therefore, we count the frequencies of words in the CDV occurring in all definitions and empirically remove the most frequent $1$\% and the infrequent $10$\%.
So far we have obtained the sememe set that is composed of $2,046$ sememes.

\subsection{Extracting Sememes from Definitions}

Next, we extract sememes for each sense of each word from its definition.
We take the word ``beautiful'' as a running example to illustrate the process of sememe extraction, as shown in Figure \ref{fig:method}.

``beautiful'' has two senses in LDOCE, and both of them are adjective.
For each sense, we first use NLTK to normalize its definition including tokenization and lemmatization.
For example, the definition of its first sense is normalized into 
a sequence of tokens: \{``someone'', ``or'', ``something'', ``that'', ``be'', ``beautiful'', ``be'', ``extremely'', ``attractive'', ``to'', ``look'', ``at''\}.
Then we remove the tokens that are not in the sememe set.
In the above example, ``someone'', ``something'', ``or'', ``that'', ``be'', ``to'' and ``at'' are removed.
So far we obtain the sememes of the first sense of ``beautiful'': \{\texttt{beautiful}, \texttt{extremely}, \texttt{attractive}, \texttt{look}\}.
In a similar way, we can obtain the sememes of its second sense: \{\texttt{good}, \texttt{give}, \texttt{pleasure}\}.

By repeating this process on all the words of LDOCE, we obtain an English SKB that we name EDSKB. 
Its statistics are shown in Table \ref{tab:stat}.

\begin{table}[!t]
\centering
\resizebox{1.02\columnwidth}{!}{
    \begin{tabular}{ccccc}
    \toprule
    SKB   & \#Word/Phrase & \#Sense & \#Sememe & \#AvgSem \\
    \midrule
    HowNet & 50,879 & 111,519 & 2,187 & 2.26  \\
    EDSKB & {70,218} & {105,160} &2,046 & 6.03  \\
    \ \ EDSKB$^*$  & {70,218} & {105,160} & \ 1,682\protect\footnote{.} & 2.04  \\
    \bottomrule
    \end{tabular}
}
\caption{Statistics of EDSKB, its distilled version EDSKB$^*$ and the English part of HowNet.\protect\footnote{.}   
\#AvgSem denotes the average sememe number per sense.}
\label{tab:stat}
\end{table}
\addtocounter{footnote}{-1}
\footnotetext{Notice that its sememe set shrinks because some sememes are not annotated to any senses anymore.}
\addtocounter{footnote}{1}
\footnotetext{The data of HowNet are obtained from OpenHowNet \citep{qi2019openhownet}.}

\subsection{Distilling Sememes of Senses}

By comparison with HowNet, we find that the sememe set of EDSKB is smaller (EDSKB $2,046$ vs. HowNet $2,187$), but its average sememe number per sense is much larger (EDSKB $6.03$ vs. HowNet $2.26$), which means the sememes of EDSKB are utilized more fully and effectively. 
Moreover, 
annotating a sense with more sememes can explain the sense more accurately and finely.
Nevertheless, it would also increase the distinguishability between different senses/words, which has an adverse effect on some downstream tasks. 
For example, word-level textual adversarial attacking conducts word substitution to generate adversarial examples,
in which fewer substitute words usually lead to lower attack success rates \citep{wang2019natural,zang2020word}.
In a sememe-based word substitution strategy \citep{zang2020word}, more sememes per sense mean fewer substitute words that share the same sememes can be found, which will decrease the final adversarial attack success rate.
To address this problem, we intend to craft an extra distilled version of EDSKB by distilling its sememes of senses.

To this end, we need to determine the importance of each sememe of a sense, and remove the relatively unimportant sememes. 
Here we resort to dependency parsing \citep{kubler2009dependency}.
Dependency parsing is used to analyze syntactic structures of a sentence by identifying the word that another word is ``dependent'' on,
e.g., 
the adjective is dependent on the noun in an adjective-noun phrase. 
We believe that the words with more dependents are more important in a definition.
Hence, we define the importance score of a sememe for a sense as the number of the dependents of its original word in the definition.  

Then, we empirically remove the sememes whose importance scores are below the highest importance score minus $t$ for the senses having $m$ or more sememes. 
Here $t$ and $m$ are two hyper-parameters and are tuned to $1$ and $4$ respectively, based on the performance on the validation sets of downstream tasks, especially adversarial attacking.  
For example, for the first sense of ``beautiful'', by using AllenNLP \citep{gardner2018allennlp} to conduct dependency parsing on its definition, we obtain the numbers of dependents of all the words in the definition.
Correspondingly, we get the importance scores of the four sememes \{\texttt{beautiful}, \texttt{extremely}, \texttt{attractive}, \texttt{look}\}, which are 2, 0, 6 and 0 respectively.
The highest importance score is 6 and thus the sememes whose importance scores are less than 5 are removed, i.e., \texttt{beautiful}, \texttt{extremely} and \texttt{look}.
Finally, the remaining sememe of the first sense of ``beautiful'' is \{\texttt{attractive}\}. 
As for the second sense, it has only 3 sememes and all of them are retained (the sememe number threshold for sememe reduction is $m=4$).
Therefore, its final sememes after reduction are still \{\texttt{good}, \texttt{give}, \texttt{pleasure}\}.

By repeating the above process on all the words, we obtain a distilled version of EDSKB (signified by EDSKB$^*$), whose average sememe number per sense is comparable with HowNet ($2.04$ vs. $2.26$).
Its detailed statistics are also shown in Table \ref{tab:stat}.

Later experiments (on both English and French) show that the full version outperforms the distilled version in some downstream tasks while not in others.
In practice, we can build both full and distilled versions and conduct experiments to see which one is better in a specific task. It is affordable to build and evaluate two versions.

\section{Intrinsic Evaluation} 
In this section, we conduct an intrinsic evaluation to assess the sememe annotation consistency of EDSKB.
Sememe annotation consistency measures how compatible the sememe annotations for different words/senses are, e.g., whether two synonyms are annotated with exactly the same sememes. 
The sememe annotation consistency of an SKB not only reflects its intrinsic quality but also has impact on its effectiveness in downstream tasks.

We evaluate both full and distilled versions of EDSKB, and the English part of HowNet for comparison.
We adopt a sememe consistency assessment method named CCSA \citep{liu2020research}, which is designed for HowNet originally but can be used for any SKB.
This method is motivated by the idea that semantically close senses should have similar sememes, which conforms to the linguistic definition of sememes.
It actually implements a sememe prediction process that predicts sememes for a small proportion of senses according to the sememe annotations of the other senses. 
The sememe prediction method it adopts is based on collaborative filtering \citep{xie2017lexical}, and tends to predict the sememes that are annotated to semantically close senses to the target sense.
Therefore, higher sememe prediction performance means the semantically close senses are annotated with more similar sememes, and the sememe annotations are more consistent.
Correspondingly, the sememe annotation consistency of an SKB is measured by two sememe prediction evaluation metrics, namely mean average precision (MAP) and F1 score.


Table \ref{tab:consistency} lists the evaluation results of sememe annotation consistency.
We can see that the distilled version of EDSKB has overall higher consistency than HowNet, and the full version of EDSKB yields lower consistency results.
It is not strange because CCSA is based on sememe prediction and according to previous work \citep{qi2020towards}, senses with more sememes usually have lower prediction performance.
Since the full version of EDSKB has much more sememes per sense than HowNet, it is actually not fair to compare their consistency using CCSA.
The distilled version of EDSKB has a similar average sememe number as HowNet, and its superior results can demonstrate the great consistency of the dictionary-based SKB. 

\begin{table}[!t]
\centering
\resizebox{.55\columnwidth}{!}{
    \begin{tabular}{ccccc}
    \toprule
    SKB   & MAP & F1   \\
    \midrule
    HowNet & {0.93} & \underline{0.91}  \\
    EDSKB & {0.88} & {0.86}  \\
    \ \ EDSKB$^*$ & \textbf{0.95} & \underline{0.91}  \\
    \bottomrule
    \end{tabular}
}
\caption{Sememe annotation consistency results. The \textbf{boldfaced} results show statistically significant improvement over the best results from baselines with \textit{p}$<$0.1 given by \textit{t}-test, and the \underline{underlined} results represent having no significant difference.\footnote{The same is true for the following tables.}}
\label{tab:consistency}
\end{table}
\footnotetext{The same is true for the following tables.}

\section{Extrinsic Evaluation}
In this section, we conduct extrinsic evaluations to assess the effectiveness of EDSKB in downstream tasks.
We pick three representative sememe-incorporated neural network models that are used for language modeling, sequence modeling and textual adversarial attacking tasks, respectively.
All of them are originally designed for HowNet and have demonstrated efficacy on their respective tasks.

\subsection{Language Modeling}
In this subsection, we try to apply EDSKB to the task of language modeling.
We use SDLM \citep{gu2018language}, a sememe incorporation method for language models, to incorporate EDSKB into two representative language models based on recurrent neural networks (RNNs).

Language modeling is aimed at predicting the next word given previous context \citep{bengio2003neural}.
Language models based on RNNs, especially LSTMs \citep{hochreiter1997long}, are very popular, which use RNNs to encode the previous text into a vector and then feed the vector to a classifier to predict the next word.  
SDLM reforms the prediction process.
Instead of directly predicting the next word, SDLM predicts sememes first, then senses and finally the next word. 

\paragraph{Base Models}
We use two representative LSTM-based language models as the base models into which EDSKB is incorporated by SDLM.

\begin{itemize} [topsep=1pt, partopsep=1pt, leftmargin=12pt, itemsep=-2pt]
	\item \textbf{Tied LSTM} \citep{zaremba2014recurrent}, which enhances a vanilla two-layer LSTM language model by introducing dropout and weight tying.
We use its large version whose word embedding and hidden vector sizes are $1,500$.

	\item \textbf{AWD-LSTM} \citep{merity2018regularizing}, 
which adopts several regularization and optimization strategies including DropConnect \citep{wan2013regularization} and non-monotonically triggered average stochastic gradient descent, and is a very strong baseline language model.
Its hidden vector size is $1,150$ and word embedding size is $400$.
\end{itemize}

\paragraph{Baseline Methods}
In addition to the two original base models, we additionally use SDLM to incorporate HowNet into the base models as baseline methods.


\paragraph{Datasets} 
We choose two benchmark language modeling datasets for evaluation, namely Penn Treebank (PTB) \citep{marcus1993building} and WikiText-2 \citep{merity2017pointer}.
PTB consists of news stories from the Wall Street Journal. 
The numbers of tokens in its training, validation and test sets are $887,521$, $70,390$ and $78,669$, respectively.
WikiText-2 is made up of Wikipedia articles, and 
it has $2,088,628$, $217,646$ and $245,569$ tokens in its training, validation and test sets.


\paragraph{Experimental Settings} 
In our experiments, we use the official implementation of SDLM
and its default hyper-parameters as well as training methods.
The evaluation metric is perplexity.
The lower perplexity a language model computes, the better the language model is. 

\paragraph{Experimental Results}
Table \ref{tab:lm} lists the perplexity results on the two datasets.
We observe that the models incorporated with EDSKB, especially the full version, consistently outperform the two base models without sememe incorporation and even the HowNet-incorporated models. 
These results demonstrate the effectiveness of EDSKB in language modeling.

\begin{table}[!t]
\centering
\resizebox{.9\columnwidth}{!}{
\begin{tabular}{lcccc}
    \toprule
    \multicolumn{1}{c}{Dataset} & \multicolumn{2}{c}{PTB}&\multicolumn{2}{c}{WikiText-2} \\
    \cmidrule{1-5} 
    \multicolumn{1}{c}{Model} & {Valid} & {Test} & {Valid} & {Test} \\
    \midrule
    Tied LSTM &  63.92 & 63.98 & 53.10 & 51.41 \\
    \quad+HowNet & 58.93 & 58.95 &  48.83 & 47.28  \\
    \quad+EDSKB & \textbf{58.81} & \textbf{58.82} &  \textbf{43.38} & \textbf{42.15}  \\
    \quad+EDSKB$^*$ & {60.17} & {60.15} &  {45.18} & {42.59}  \\
    \midrule
    AWD-LSTM & 58.89 & 59.24 & 45.29 & 44.13   \\
    \quad+HowNet & 58.95 & 58.92 & 46.84 & 45.29   \\
    \quad+EDSKB  & \textbf{56.94} & \textbf{57.13} & \textbf{42.44} & \textbf{41.25}  \\
    \quad+EDSKB$^*$ & {58.63} & {58.59} &  {43.85} & {43.95}  \\
    \bottomrule
\end{tabular}
}
\caption{Perplexity results of different language models on the validation and test sets of PTB and WikiTex-2.}
\label{tab:lm}
\end{table}

\subsection{Sequence Modeling}
In this subsection, we incorporate EDSKB into RNNs to improve their sequence modeling ability by
SememeCell \citep{qin2020improving}, a sememe incorporation method for enhancing RNNs.

SememeCell uses a special RNN cell to encode sememes of a word into a latent vector and transmits it to the corresponding RNN cell of the word, aiming to inject the semantic information of sememes into RNNs. 
It has demonstrated its effectiveness in improving the sequence modeling ability of RNNs in multiple downstream tasks, including natural language inference, sentiment analysis and paraphrase detection \citep{qin2020improving}.


\paragraph{Base Models}
Following \citet{qin2020improving}, we choose two most representative RNNs, namely LSTM, GRU \citep{cho2014learning}, and their bidirectional versions (BiLSTM and BiGRU) as the base models, into which sememes are incorporated by SememeCell.

\paragraph{Baseline Methods}
In addition to the vanilla and HowNet-incorporated RNNs, we also design another two baseline methods.

\begin{itemize} [topsep=1pt, partopsep=-15pt, leftmargin=12pt, itemsep=-2pt]
\item \textbf{+Pseudo}. RNNs incorporated with either EDSKB or HowNet have a little more parameters than vanilla RNNs.
	To eliminate the possible effect brought by more parameters, we build a pseudo-SKB named Pseudo.
	Specifically, for each sense in EDSKB, we substitute its sememes with the same number of meaningless labels.
	The labels are randomly sampled from a label set with the same size as the sememe set of EDSKB.
	We use SememeCell to incorporate this pseudo-SKB into the two base models as baselines, which have exactly the same numbers of parameters as EDSKB-incorporated models.

\item \textbf{+Definition}. EDSKB is obtained from dictionary definitions by the transformation from a sequence of words (definition) into several discrete semantic labels (sememes).
We intend to compare the EDSKB-incorporated models and models incorporated with the complete dictionary definitions.
Since SememeCell only takes a vector (i.e., the sum of sememe embeddings) as input, we can leverage it to incorporate definitions into RNNs by encoding definitions into vectors with a sentence encoder. 
Specifically, we choose the powerful pre-trained language model BERT \citep{devlin2019bert} as the sentence encoder and use the hidden vector of the \texttt{[CLS]} token as the definition embedding.
The definition-incorporated RNN models are also baselines. 
\end{itemize}

\paragraph{Downstream Tasks and Datasets}
RNNs are basic sequence encoders and can be used in many downstream NLP tasks.
Following \citet{qin2020improving}, we choose two representative tasks to evaluate the sentence modeling ability of EDSKB-incorporated RNNs.
\begin{itemize} [topsep=1pt, partopsep=1pt, leftmargin=12pt, itemsep=-2pt]
\item Natural language inference (NLI), which is aimed at determining whether a natural language hypothesis can be inferred from a premise. It is a typical sentence pair classification task. We use the SNLI dataset \citep{bowman2015alarge} for evaluation. SNLI contains about $570,000$ English premise-hypothesis pairs, and each pair is manually labeled one of three relation labels, namely ``entailment'', ``contradiction'' and ``neutral''.

\item Sentiment analysis, which aims to recognize the sentiment orientation of a sentence and is a typical single sentence classification task. 
	Following \citet{qin2020improving}, we use the CR dataset \citep{hu2004mining} for evaluation. 
	It contains about $8,000$ product reviews and each review is labeled with ``positive'' or ``negative''.
\end{itemize}

\paragraph{Experimental Settings}
We use the official implementation of SememeCell \citep{qin2020improving} and the default hyper-parameter settings and training methods, where the embedding size (for both word and sememe embeddings) is $300$ and hidden size is $2,048$.
In the baseline method +Definition, to keep the definition vector size comparable with sememe embedding size, we choose the medium version of BERT, which has $512$-dimensional hidden vectors and $8$ layers.\footnote{\url{https://github.com/google-research/bert}}
As for evaluation metrics, we use accuracy for both NLI and sentiment analysis.

\paragraph{Experimental Results}
Table \ref{tab:sl} shows the evaluation results on the test sets of SNLI and CR.
We can see that RNNs incorporated with dictionary-based SKB, especially the full version (+EDSKB), yield overall better results than vanilla RNNs, which proves that the dictionary-based SKB can improve the sequence modeling ability of RNNs. 
Furthermore, the +EDSKB models
outperform +Pseudo models that have the same number of parameters, +Definition models that have the same semantic information source, and +HowNet models that incorporate another SKB.
These results demonstrate the superiority of discrete sememes over definitions, and the advantage of dictionary-based SKB over HowNet in enhancing RNNs. 
+Pseudo performs slightly better than vanilla in some cases, which is probably because +Pseudo utilizes the random meaningless labels as noises. 
The addition of noise has been proven a regularization method for mitigating overfitting and improving performance in neural networks \citep{bishop1995training}.

\begin{table}[!t]
\centering
\resizebox{1.02\linewidth}{!}{
\begin{tabular}{cl|cccc}
    \toprule
    Dataset & \multicolumn{1}{c|}{Method} & {LSTM} & {GRU} & {BiLSTM} & {BiGRU} \\
    \midrule
\multirow{6}{*}{SNLI} & vanilla & 80.66 & 82.00 & 81.30 & 81.61 \\
    & \ +Pseudo & 81.28 & 80.90 & 81.91 & 82.07 \\
    & \ +HowNet & 81.87 & {82.90} & \underline{82.55} & {83.15} \\
    & \ +Definition & 81.62 & 82.80 & 81.10 & 83.22 \\
    & \ +EDSKB & \textbf{82.32} & \textbf{83.18} & \underline{82.54} & \textbf{83.55}  \\
    & \ +EDSKB$^*$ & 81.78 & 82.10 & 82.11 & 82.35 \\
    \midrule
\multirow{6}{*}{CR} & vanilla &  74.17 & 76.37 & 77.62 & 78.76 \\
    & \ +Pseudo & 73.96 & 75.44 & 76.16 & 78.20  \\
    & \ +HowNet & 76.47 & 78.57 & 77.66 & 76.25  \\
    & \ +Definition & 76.29 & 78.20 & 77.19 & 77.77 \\
    & \ +EDSKB & \textbf{77.51} & \textbf{79.68} & \textbf{78.95} & \textbf{78.88}  \\
    & \ +EDSKB$^*$ & 75.09 & 77.54 & 76.90 & 78.18 \\
     \bottomrule
\end{tabular}
}
\caption{Accuracy results of different models on the test sets of SNLI and CR.}
\label{tab:sl}
\end{table}

\subsection{Textual Adversarial Attacking}
\label{sec:attack}
In this subsection, we investigate the effectiveness of EDSKB in textual adversarial attacking.

Adversarial attacking has attracted considerable research attention recently, mainly because it can reveal the vulnerability of neural network models and help improve their robustness and interpretability \citep{xu2020adversarial}. 
Adversarial attacks use \textit{adversarial examples} \citep{szegedy2014intriguing}, which are maliciously crafted by perturbing the original model input, to fool the victim model.
In textual adversarial attacking, 
word-level attack methods, mainly based on word substitution, are a kind of popular attack method and have demonstrated overall better attack performance \citep{wang2019natural}.

\citet{zang2020word} decompose the process of word-level attacks into two steps:
(1) determining the substitute set for each word in the original input via a word substitution strategy, e.g., synonym-based and word embedding-based substitution strategies;
and (2) searching the combinations of each original word's substitutes for adversarial examples that can successfully fool the victim model.

They also propose an adversarial attack approach that employs a sememe-based word substitution strategy and achieves strong attack performance.
The sememe-based word substitution strategy essentially regards a word $w_1$ as the substitute of another word $w_2$, if one sense of $w_1$ has the same sememes as one sense of $w_2$, according to an SKB.
We use this approach to conduct textual adversarial attacks and measure the attack performance.

\paragraph{Baseline Methods}
In addition to the original sememe-based attack approach that uses HowNet as the SKB, we choose some other baseline methods for comparison.
Notice that all these baseline methods use the same approach to search for adversarial examples (the aforementioned step 2) and differ in word substitution strategies (step 1) only. 

\begin{itemize} [topsep=1pt, partopsep=1pt, leftmargin=12pt, itemsep=-2pt]
\item \textbf{+Synonym}, the attack method that uses synonym-based word substitution strategy. Following previous work \citep{ren2019generating}, we use WordNet as the thesaurus and the words in a synset can be regarded as substitutes of each other.  

\item \textbf{+Definition}, the attack method that uses a definition-based word substitution strategy. Inspired by word embedding-based word substitution, we encode the definition of each sense of words into a vector and define the similarity between two words as the cosine similarity between their closest definition vectors. 
    Then, a certain number of words that are most similar to the target word are regarded as its substitutes.
    Specifically, we still use the medium-size BERT to encode definitions into $512$-dimensional vectors.
    And the number of substitutes of each word is the same as that in the sememe-based substitution strategy.
\end{itemize}
In this task, the +Pseudo baseline in the previous section cannot work because it would regard random words as substitutes of the target word.

\paragraph{Victim Models and Datasets}
Following \citet{zang2020word}, 
we choose BiLSTM and BERT, specifically BERT$_{\rm BASE}$ as the victim models we intend to attack.
The evaluation task is sentiment analysis and the evaluation dataset is SST-2 \citep{socher2013recursive}.
SST-2 comprises about $10,000$ sentences in movie reviews and each sentence is labeled with ``positive'' or ``negative''.
The accuracy results of BiLSTM and BERT on the test set of SST-2 are $83.75$ and $90.28$.

\paragraph{Experimental Settings}
We use the official implementation of the sememe-based attack approach \citep{zang2020word} and the default hyper-parameter settings. 

\paragraph{Evaluation Metrics}
Following \citet{zang2020word}, we use attack success rate to measure the effectiveness of an attack method and three metrics to assess the quality of its adversarial examples.
The three metrics are (1) word modification rate, the percentage of words in the original input that are perturbed; (2) increase rate of grammatical errors in adversarial examples compared with original input, where LanguageTool grammar checker is used; and (3) perplexity given by GPT-2 \citep{radford2019language} that is used to measure the fluency of adversarial examples.
The lower the three metrics are, the better the quality of adversarial examples is.

\begin{table}[!t]
\centering
\resizebox{\columnwidth}{!}{
\begin{tabular}{cl|cccc}
    \toprule
    Victim & Attack Method & ASR & \%M & \%IGE & PPL  \\ 
    \midrule
\multirow{5}{*}{BiLSTM} & +Synonym & 79.0 & 10.45 & 7.59 & 593.09 \\
    & +Definition & 90.0 & 8.76 & 7.56 & 518.71 \\
    & +HowNet & 93.6 & 9.02 & 2.57 & \textbf{468.92} \\
    & +EDSKB & 26.5 & \underline{8.27} & 3.77 & 538.46  \\
    & +EDSKB$^*$ & \textbf{94.0} & \underline{8.29} & \textbf{1.27} & 507.34  \\
    \midrule
\multirow{5}{*}{BERT} & +Synonym & 81.3 & 9.22 & 8.00 & 576.82 \\
    & +Definition & 86.3 & 8.03 & 7.18 & 538.00 \\
    & +HowNet & 91.2 & 8.25 & 2.08 & {503.06} \\
    & +EDSKB & 29.7 & 8.10 & 3.36 & \textbf{485.00}  \\
    & +EDSKB$^*$ & \textbf{93.3} & \textbf{7.66} & \textbf{1.07} & 544.51  \\     
    \bottomrule
\end{tabular}
}
\caption{Adversarial attack results of different word substitution strategies. ASR is short for attack success rate. \%M, \%IGE and PPL denote word modification rate, increase rate of grammatical errors and perplexity, respectively.}
\label{tab:attack}
\end{table}

\begin{table*}
\centering
\resizebox{\linewidth}{!}{
\begin{tabular}{c|c|l}
    \toprule
   	Word & SKB & \multicolumn{1}{c}{Sememes} \\
   	\midrule
   	\multirow{3}{*}{screenwriter} & HowNet & \texttt{human}, \texttt{occupation}, \texttt{entertainment}, \texttt{compile}, \texttt{shows} \\
   	\cline{2-3}
   	& EDSKB & \texttt{someone}, \texttt{write}, \texttt{play}, \texttt{film}, \texttt{television} \\
   	\cline{2-3}
   	& \ \ EDSKB$^*$ & \texttt{write}, \texttt{play}, \texttt{film}, \texttt{television} \\
   	\midrule
   	
   	\multirow{5}{*}{tweet} & {HowNet} & \texttt{InstitutePlace}, \texttt{ProperName}, \texttt{produce}, \texttt{software}, \texttt{LookFor}, \texttt{document}, \texttt{information}, \texttt{internet}\\
   	\cline{2-3}
   	& \multirow{2}{*}{EDSKB} & Sense 1: \texttt{bird}, \texttt{make}, \texttt{high}, \texttt{small}, \texttt{short}, \texttt{sound} \\
   	& & Sense 2: \texttt{service}, \texttt{message}, \texttt{network}, \texttt{short}, \texttt{send}, \texttt{use}, \texttt{social} \\
   	\cline{2-3}
   	& \multirow{2}{*}{\ \ EDSKB$^*$} & Sense 1: \texttt{bird}, \texttt{sound} \\
   	& &  Sense 2: \texttt{message}, \texttt{send}, \texttt{use}, \texttt{network} \\
   	
	\bottomrule
\end{tabular}
}
\caption{Two cases of sememe annotations in HowNet, EDSKB and EDSKB$^*$.}
\label{tab:case}
\end{table*}

\paragraph{Experimental Results}
According to Table \ref{tab:attack},
we find that the attack method based on EDSKB$^*$ not only achieves the highest attack success rates but also generates adversarial examples with overall higher quality. 
These results show that the dictionary-based SKB EDSKB$^*$ can better capture the semantic relations between words and find appropriate substitutes for adversarial attacks.
Attack success rates of the EDSKB-based method are extremely low.
It is because EDSKB has too many sememes per sense, which causes the found substitutes to be very few (EDSKB 1.6, EDSKB$^*$ 12.6 and HowNet 15.3 on average), according to the sememe-based word substitution strategy that requires substitutes to have the same sememes.

\section{Case Study on Sememe Annotations}
In this section, we give two cases of sememe annotations in EDSKB and EDSKB$^*$ as well as HowNet in Table \ref{tab:case}.

The first case is the word ``screenwriter''.
In HowNet, this word has only one sense that is annotated by five sememes, as listed in the second row of Table \ref{tab:case}.
As for EDSKB and EDSKB$^*$, according to \textit{Longman Dictionary of Contemporary English} (LDOCE), this word also has only one sense whose definition is ``someone who writes plays for film or television''.
EDSKB provides five sememes and one (\texttt{someone}) is filtered out in EDSKB$^*$.
By comparison, we can find that sememes in EDSKB and EDSKB$^*$ can represent the meaning of the word more specifically, e.g, \texttt{write} and \texttt{play}, while sememes in HowNet seem to express a more general meaning.

The second case is about the word ``tweet''.
HowNet only annotates one sense for this word, i.e., ``to send a message on Twitter''.
As for EDSKB and EDSKB$^*$, since LDOCE contains the basic meaning of this word, i.e., ``to make the short high sound of a small bird'', the sememes including \texttt{bird} and \texttt{sound} are extracted to express this meaning.
In addition, for the shared sense, sememes in EDSKB and EDSKB$^*$ are more succinct than those in HowNet, e.g., \texttt{message} in EDSKB/EDSKB$^*$ can better describe the core meaning of ``tweet'' than \texttt{document} and \texttt{information} in HowNet.

From the two cases, we can see the advantage of the dictionary-based SKBs over HowNet in terms of sememe annotations.
We hope that the dictionary-based SKBs can be used to perfect HowNet by supplying more senses and annotating more suitable sememes.

\vspace{-3pt}
\section{Conclusion and Future Work}
\vspace{-3pt}
In this paper, we propose to utilize a dictionary to build an SKB for the first time, which can be implemented by a simple, quick and fully automatic process.
We try utilizing existing dictionaries to build an English SKB and a French SKB, and demonstrate their effectiveness on multiple NLP tasks.
Extensive experimental results prove the reliability and practicality of our idea about dictionary-based SKB construction.



It is worth mentioning that although EDSKB delivers better empirical results than HowNet, HowNet has its unique advantages including better interpretability and multilinguality. 
In the future, therefore, we will systematically compare the sememe annotations in EDSKB and HowNet and try to use EDSKB to improve and expand HowNet.
Besides, the hierarchical structures of sememes in HowNet are neglected in this paper. 
We will also explore to extract sememes with hierarchy from dictionary definitions.

\section*{Acknowledgements}
This work is supported by the National Key Research and Development Program of China (Grant No. 2020AAA0106502 and No. 2020AAA0106501) and Beijing Academy of Artificial Intelligence (BAAI).
We also thank all the anonymous reviewers for their valuable comments and suggestions.

\section*{Ethical Considerations}
In this paper, we use two copyrighted dictionaries, namely Longman Dictionary of Contemporary English and Le Petit Robert French Dictionary.
We extract data from the electronic versions of the two dictionaries we bought for the research purpose only.
We will not release the dictionary data.
In addition, the datasets we use in downstream tasks are all open and free (actually also widely used).

The task we tackle is sememe knowledge base construction, which is not a practical application and is only related to NLP research.
Therefore, the datasets we build and the models we use would not be misused by common people.

In addition, since we do not use very large models, the required energy in this work is very limited.
Finally, we use no demographic or identity characteristics.

\bibliographystyle{acl_natbib}
\bibliography{acl2021}

\appendix
\section{Building Process of a French SKB}
\label{sec:french}
In this section, we describe the building process of a dictionary-based French SKB that we call FDSKB.
We choose Le Petit Robert French Dictionary 2016 edition \citep{Robert2016Dictionnaire}, a very popular French dictionary, as the base dictionary.

\paragraph{Constructing the Sememe Set}
We first construct a sememe set from the defining vocabulary of the dictionary.
Similar to EDSKB, we remove the most frequent and infrequent words that appear in definitions as well as some stop words, and obtain a sememe set comprising $2,919$ sememes (defining words).

\paragraph{Extracting Sememes from Definitions}
We use Stanza \citep{qi2020stanza} to tokenize and lemmatize the definitions of all words in the dictionary and extract the sememes of each sense of each word according to the sememe set.
So far, we have obtained the full version of FDSKB, whose statistics are shown in Table \ref{tab:french-stat}.

\paragraph{Distilling Sememes of Senses}
We adopt a similar way to EDSKB to distill the sememes of senses.
Specifically, we use Stanza to conduct dependency parsing for every definition and obtain the importance score of each sememe.
Then we empirically remove the unimportant sememes according to the experimental results of downstream tasks.
In this way, we obtain the distilled version of FDSKB (FDSKB$^*$), whose statistics are also in Table \ref{tab:french-stat}.

\section{Evaluation of FDSKB}
\label{sec:french-eval}
In this section, similar to EDSKB, we conduct both intrinsic and extrinsic evaluations for FDSKB and FDSKB$^*$.
Notice that since HowNet covers only English and Chinese, there are no available HowNet-based baseline methods for French.

\subsection{Intrinsic Evaluation}
We still use CCSA \citep{liu2020research} to measure the sememe annotation consistency.
The MAP and F1 score for FDSKB are $83.47$ and $80.51$ respectively, and those for FDSKB$^*$ are $90.03$ and $90.01$ respectively.
These results are comparable to those of EDSKB and can prove good sememe annotation consistency of FDSKB.
Besides, similar to EDSKB, the distilled version FDSKB$^*$ delivers better sememe annotation consistency than FDSKB because it has fewer sememes per sense.

\subsection{Extrinsic Evaluation}
We conduct extrinsic evaluation for FDSKB and FDSKB$^*$ on three tasks including language modeling, natural language inference (NLI) and text classification.

\subsubsection*{Language Modeling}
Similar to EDSKB, we use SDLM \citep{gu2018language} to incorporate FDSKB into Tied LSTM \citep{zaremba2014recurrent} and AWD-LSTM \citep{merity2018regularizing}.
The experimental settings are the same as those in English experiments.

\begin{table}[!t]
\centering
\resizebox{\columnwidth}{!}{
    \begin{tabular}{ccccc}
    \toprule
    SKB   & \#Word/Phrase & \#Sense & \#Sememe & \#AvgSem \\
    \midrule
    FDSKB & 55,836 & 113,722    & 2,919 & 4.32   \\
    \ \ FDSKB$^*$  & 55,836 & 113,722  & 2,919 & 1.97  \\
    \bottomrule
    \end{tabular}
}
\caption{Statistics of FDSKB and its distilled version FDSKB$^*$. 
\#AvgSem denotes the average sememe number per sense.}
\label{tab:french-stat}
\end{table}

\begin{table}[!t]
\centering
\resizebox{.9\columnwidth}{!}{
\begin{tabular}{lcccc}
    \toprule
    \multicolumn{1}{c}{Dataset} & \multicolumn{2}{c}{French News}&\multicolumn{2}{c}{FR-Wikipedia} \\
    \cmidrule{1-5} 
    \multicolumn{1}{c}{Model} & {Valid} & {Test} & {Valid} & {Test} \\
    \midrule
    Tied LSTM &  17.02 & 18.35 & 17.23 & 15.75 \\
    \quad+FDSKB & \underline{15.72} &  \underline{16.85} & 17.14 & 15.60 \\
    \quad+FDSKB$^*$ & \underline{15.70} & \underline{16.89} & \textbf{16.50} & \textbf{15.15} \\
    \midrule
    AWD-LSTM & 18.41 & 19.71 & 16.76 & 15.30 \\
    \quad+FDSKB  & 15.41 & 16.47 & \textbf{15.45} & 15.72 \\
    \quad+FDSKB$^*$ & \textbf{14.03} & \textbf{15.14} & 15.90 & \textbf{14.10} \\
    \bottomrule
\end{tabular}
}
\caption{Perplexity results on the validation and test sets of French News and FR-Wikipedia. The \textbf{boldfaced} results show statistically significant improvement over the best results from baselines with \textit{p}$<$0.1 given by \textit{t}-test, and the \underline{underlined} results represent having no significant difference.\footnote{The same is true for the following tables.}}
\label{tab:french-lm}
\end{table}

We choose two evaluation datasets: 
(1) \textbf{French News}\footnote{\url{https://webhose.io/free-datasets/french-news-articles/}}, which comprises French news articles from popular news sites. It has $2,131,774$ / $358,972$ / $370,059$ tokens in its training / validation / test sets.
(2) \textbf{FR-Wikipedia}\footnote{\url{http://redac.univ-tlse2.fr/corpora/wikipedia_en.html}}, which is composed of French Wikipedia articles.
The token numbers in its training / validation / test sets are $3,252,094$ / $520,333$ / $517,669$.

The experimental results are given in Table \ref{tab:french-lm}.
We observe that both FDSKB and FDSKB$^*$ bring decreases of perplexity, which demonstrates the effectiveness of the French dictionary-based SKB in language modeling and the practicality of our dictionary-based SKB building method.
Notice that since the perplexity results of original Tied LSTM and AWD-LSTM on the two French datasets are quite good, the enhancement brought by FDSKB is comparatively less than that in English.

\begin{table}[!t]
\centering
\resizebox{\linewidth}{!}{
\begin{tabular}{cl|cccc}
    \toprule
    Dataset & \multicolumn{1}{c|}{Method} & {LSTM} & {GRU} & {BiLSTM} & {BiGRU} \\
    \midrule
\multirow{6}{*}{XNLI} & vanilla & 61.14 & 60.88 & 61.56 & 61.36 \\
    & \ +Pseudo & 60.96 & 61.46 & 61.10 & 61.12 \\
    & \ +Definition & 61.64 & 61.10 & 61.80 & 61.86\\
    & \ +FDSKB & 62.38 & 61.54 & \textbf{62.52} &  61.44\\
    & \ +FDSKB$^*$ & \textbf{62.52} & \textbf{61.74} & 62.16 &  \textbf{62.06}\\
    \midrule
\multirow{6}{*}{MARC} & vanilla &  79.98 & 79.16 & 80.35 & 80.64 \\
    & \ +Pseudo & 80.35 & 79.38 & 79.16 & 80.53  \\
    & \ +Definition & 81.10 & 79.39 & 80.65 & 80.80 \\
    & \ +FDSKB & 81.39 & 80.58 & 80.65 & \textbf{82.14}  \\
    & \ +FDSKB$^*$ & \textbf{81.43}  & \textbf{81.10} & \textbf{80.98} & 81.84 \\
     \bottomrule
\end{tabular}
}
\caption{Accuracy results of different models on the test sets of XNLI and MARC.}
\label{tab:french-cls}
\end{table}

\subsubsection*{NLI and Text Classification}
Similar to EDSKB, we use SememeCell \citep{qin2020improving} to incorporate FDSKB into RNNs and measure the improvement of sequence modeling ability on the tasks of NLI and text classification.

The base models are still LSTM, GRU, BiLSTM and BiGRU.
And the baseline methods are also +Pseudo and +Definition.
Here we use FlauBERT \citep{le2020flaubert}, a French pre-trained language model, to encode definitions into $768$-dimensional vectors.
The other experimental settings are the same as English.

As for evaluation datasets, we use XNLI \citep{conneau2018xnli} and MARC \citep{keung2020multilingual} respectively.
XNLI is a cross-lingual NLI dataset in 15 languages.
It is based on another NLI dataset MNLI \citep{williams2018broad} and constructs its training set by machine translation, which has $361,469$ sentence pairs.
It has $2,500$ and $5,000$ sentence pairs in the validation and test sets which are manually translated from English.
MARC (Multilingual Amazon Reviews Corpus) is a large corpus of Amazon reviews in 6 languages.
We use its French part for product category classification.
It has $40,000$ / $1,323$ / $1,345$ reviews in its  training / validation / test sets.

Table \ref{tab:french-cls} shows the accuracy results on the test sets of XNLI and MARC.
The results are basically consistent with the experimental results in English datasets.
The incorporation of the dictionary-based SKB can improve the performance of RNN models on the two different tasks, which reflects that the SKB has enhanced the sequence modeling ability of RNNs.
Moreover, the results also demonstrate the usefulness and effectiveness of our dictionary-based SKB and its building method.

\section{Experiment Running Environment}

For all the experiments, we use a server whose major configurations are as follows: (1) CPU: Intel(R) Xeon(R) E5-2680 v4 @ 2.40GHz, 56 cores; (2) RAM: 125GB; (3) GPU: 8 Nvidia RTX2080 GPUs, 12GB memory.
The operation system is Ubuntu 18.04.2 LTS (GNU/Linux 4.15.0-108-generic x86\_64).
We use PyTorch\footnote{\url{https://pytorch.org/}} v1.5.0 and Python v3.6.9 as the programming framework for the experiments on neural network models.

\end{document}